\documentclass{article}
\pdfoutput=1
\PassOptionsToPackage{numbers, compress}{natbib}
\usepackage{stackengine}
\usepackage[final]{nips_2016}
\usepackage{caption}
\usepackage{subcaption}
\usepackage[utf8]{inputenc} 
\usepackage[T1]{fontenc}    
\usepackage{hyperref}       
\usepackage{url}            
\usepackage{booktabs}       
\usepackage{amsfonts}       
\usepackage{nicefrac}       
\usepackage{microtype}      
\usepackage{graphicx}

\title{Learning Graph While Training: An Evolving Graph Convolutional Neural Network}

\author{
  Ruoyu Li \\
  Computer Science and Engineering\\
  The University of Texas at Arlington\\
  \texttt{ruoyu.li@mavs.uta.edu} \\
   \And
  Junzhou Huang \\
  Computer Science and Engineering\\
  The University of Texas at Arlington\\
  \texttt{jzhuang@uta.edu}
}

\begin{document}
\def\stackalignment{l}
\maketitle

\begin{abstract}
    Convolution Neural Networks on Graphs are important generalization and extension of classical CNNs. While previous works generally assumed that the graph structures of samples are regular with unified dimensions, in many applications, they are highly diverse or even not well defined. Under some circumstances, e.g. chemical molecular data, clustering or coarsening for simplifying the graphs is hard to be justified chemically. In this paper, we propose a more general and flexible graph convolution network (EGCN) fed by batch of arbitrarily shaped data together with their evolving graph Laplacians trained in supervised fashion. Extensive experiments have been conducted to demonstrate the superior performance in terms of both the acceleration of parameter fitting and the significantly improved prediction accuracy on multiple graph-structured datasets.
\end{abstract}

\section{Introduction}
\label{sec:intro}
Convolutional Neural Networks (CNNs) have been proven supremely successful on solving a wide variety of machine learning problems \cite{hinton2012deep}. The stationarity of data and the metric of grid unlock the possibility of designing a local convolutional kernel that linearly combines local features.  With the power of deep architecture, the network can output high-level representation of both local features and universal structures of signal. Even though the CNNs have been successful in tasks where data have underlying grid structure, e.g. text, images and videos, in many problems the data lie on irregular grid or more generally in non-Euclidean domains, e.g. molecular data, social networks and knowledge instances. Those data are better to be structured as graph, which is capable of handling varying node neighborhood connectivity and non-Euclidean metric. Under such a circumstance, the stationarity, locality and compositionality, which allow kernel-based convolution and pooling in CNNs, are no longer satisfied. The classical CNNs cannot directly work on graph-structured data.

However, a generalization of classical CNNs from regular grid to irregular graph is not straightforward. For simplicity of constructing kernel, many previous works assume data is still on low-dimensional graph and the training data has unified graph Laplacian shared across signal domain \cite{bruna2013spectral,henaff2015deep}. As a result, the graphs have to be of identical dimensions, which makes it impossible to construct an end-to-end deep learning pipeline that accepts arbitrary graph inputs. Moreover, current graph convolution layer does not deeply exploit information given by vertex connectivity due to the difficulty of designing a kernel flexible with varying neighborhood \cite{atwood2016diffusion,chen2016compressing}. Whereas, some sorts of data on non-Euclidean domain, such as molecular data, have underlying  graph structure or some prior knowledge of how to construct it, e.g. social network, many others do not have such knowledge. So, it is necessary to estimate the similarity matrix before performing graph convolutions. The state-of-the-art graph construction methods are classified into unsupervised and supervised ones \cite{henaff2015deep}. However, both graph constructions are accomplished before feeding data into the network. Therefore, the generated graph structure for the data keeps unchanged and will not be updated during the training procedure \cite{bruna2013spectral,henaff2015deep}. 

Although the supervised graph construction with fully connected networks has been exploited in DNN \cite{henaff2015deep}, their dense training weights restrict it to small graphs. Moreover, the graph structure learned from a fully connected architecture is not guaranteed to best serve the convolution neural network. To tackle these challenges, we  introduce a new graph convolution layer embedded with metric learning, so that each convolution layer is able to dynamically construct and learn graph structures for each individual data sample in the batch based on the given supervised information. Directly learning the similarity matrix has $\mathcal{O}(N^2)$ complexity for a graph of $\mathbb{R}^{N\times d}$ data. If harnessing a supervised metric learning with \emph{Mahalanobis} distance, we could reduce the parameter number to at most $\mathcal{O}(d^2)$ or even $\mathcal{O}(dm)$. As a consequence, the learning complexity becomes independent of graph size $N$. 
In classical CNNs, back-propagation generally updates kernel weights to adjust the relationship between neighboring nodes at each feature dimension individually. Then it sums up signals from all filters to construct hidden-layer activations. To grant graph CNNs a similar capability, we propose a re-parameterization on the feature dimension of graph data with additional weights and bias. 

Even for those data with inherent graph structure, it is still interesting to ask if the free graphs optimally serve the specific learning tasks based on the supervised information or not. For example, the chemical bonds, connecting a pair of atoms, directly lead to a underlying graph for each chemical compound. It is not hard to find that those chemical connections are not always the optimal information source for predicting desired outputs of specific tasks.  Consequently, it is emerging to develop new approach that automatically discovers the \emph{hidden} and \emph{task-related} graph structures that boost the performance of graph CNNs for specific task. Motivated by deep residue learning \cite{he2016deep}, we propose a residual graph Laplacian learning method, which is able to learn an optimal graph structure for each data sample and the prediction neural network simultaneously.

In this paper, we explore our approach primarily on chemical molecular datasets, although the network can be straightforwardly trained on other graph-structured data, such as point cloud, social networks and so on. Our contributions can be summarized as follows:
\begin{itemize}
\item A novel spectral graph convolution layer boosted by Laplacian learning (SGC-LL) has been proposed to dynamically update the residual graph Laplacians via metric learning for deep graph learning.
\item Re-parametrization on feature domain has been introduced in $K$-hop spectral graph convolution to enable our proposed deep graph learning and to grant graph CNNs the similar capability of feature extraction on graph data as that in the classical CNNs on grid data. 
\item An evolving graph convolution network (EGCN) has been designed to be fed by a batch of \emph{arbitrarily} shaped graph-structured data. The network is able to construct and learn for each data sample the graph structure that best serves the prediction part of network. Extensive experimental results indicate the benefits from the evolving graph structure of data.
\end{itemize}      
The rest of the paper is organized as follows. Section 2 reviews previous related works. Section 3 introduces the proposed spectral graph convolution boosted by residual Laplacian learning. Section 4 demonstrates both visual and numerical results. Section 5 concludes this paper.

\section{Related Work}
\label{sec:related}
There have been lots of works that explored local receptive fields on grid \cite{krizhevsky2012imagenet,coates2011selecting} with deep learning. However, there are not so many works on generalizing deep convolutional network to graph-structured data. The first trial of formulating CNN analogy on irregular domains modeled as graphs has been accomplished by \cite{bruna2013spectral}, who investigated performing convolution on both spatial and spectral domains of graph representations. Their works gave a spatial localized filter by designing smooth spectral kernel constructed by B-spine interpolation, but it only worked on low-dimensional  graph. 
\begin{figure}[h]
	\centering
	\includegraphics[width=0.5\textheight]{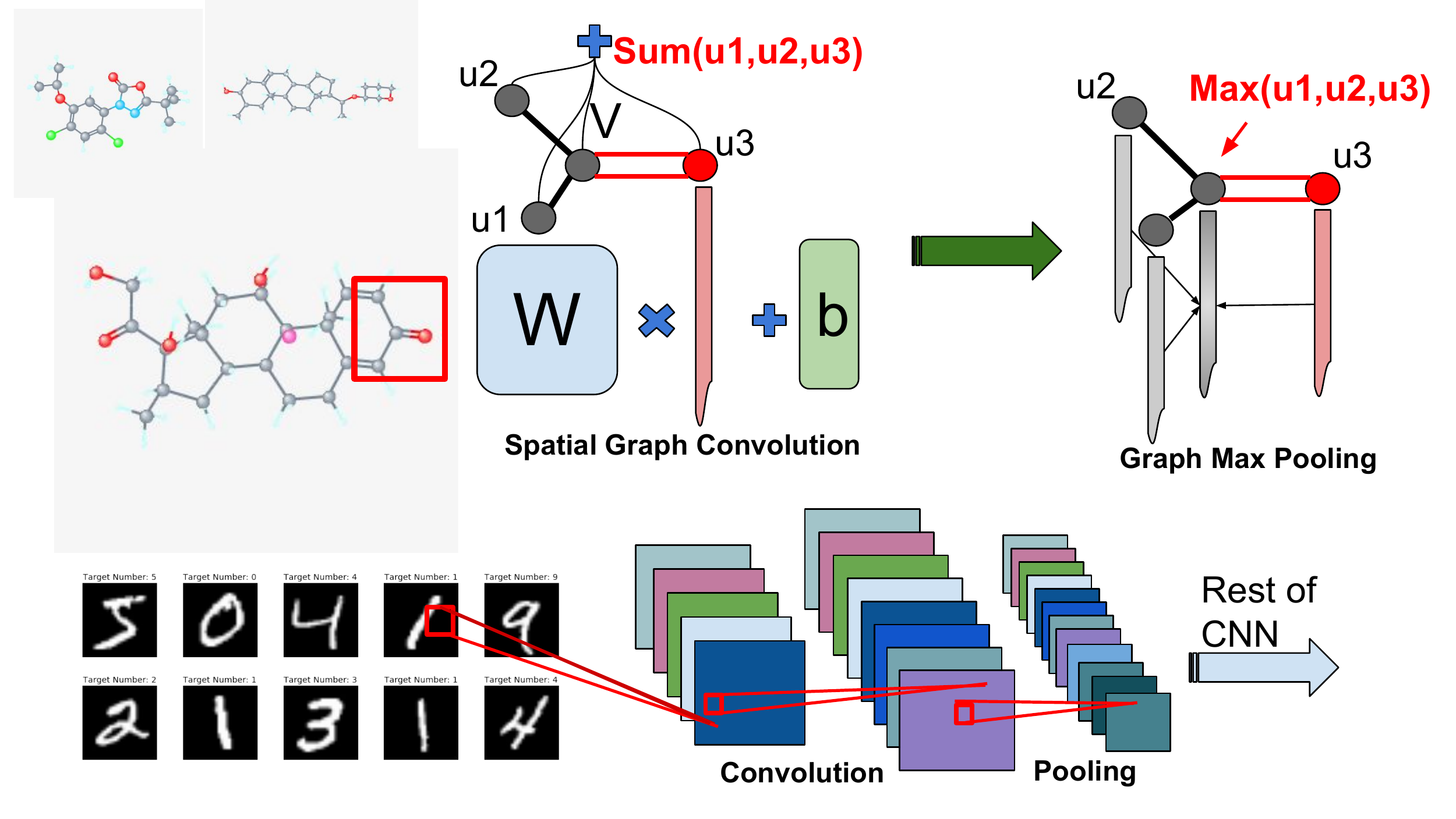}
	\caption{Network architecture of a spatial graph CNN on graphs (molecular data) compared with classical CNN on grids (images). A simplified spatial graph convolution operator sums up transformed features of neighbor nodes ($u_1, u_2, u_3$): we use separate weight $W_{u_k}$ and bias $b_{u_k}$ for neighbor $u_k$: $W_{u_k}*f_{u_k} + b_{u_k}$, $k=1,2,3$. While a graph max pooling keeps the maximum among neighbors ($u_1,u_2,u_3$) and itself node $v$ along the feature dimension. Red boxes are the convolution kernel. However, on graphs, the convolution kernel cannot work in the way same as on grid: for nodes, their neighborhood differ in both the degree and the connection type (bond).  Better viewed in color print.}\label{fig:spatial}
\end{figure}
\cite{henaff2015deep} further extended the spectral construction to a larger scale of high-dimensional graphs as well as proposed two graph construction methods in both unsupervised and supervised fashion. Inspired by previous jobs and based on graph signal processing (GSP) \cite{shuman2013emerging}, \cite{defferrard2016convolutional} introduced a new spectral graph theoretical formulation and used Chebyshev polynomials and its approximate evaluation scheme to reduce the computational cost and achieve localized filtering. \cite{kipf2016semi} showed a first-order approximation to the Chebyshev polynomials as the graph filter spectrum, which requires less training parameters.

Besides above papers on constructing convolution layer on graphs, many others studied the problem from a different angle. \cite{niepert2016learning} first investigated learning a network from a set of heterogeneous graphs to predict node-level feature as well as to do graph completion, although it is based on node sequence selection. \cite{atwood2016diffusion} introduced a graph diffusion process, which delivers equivalent effect as convolution has, but \cite{atwood2016diffusion}'s DCNN has no dependency on the indexing of nodes. Its constrains are the highly restricted locality by diffusion process and the expensive dense matrix multiplication. 

Recently, \cite{simonovsky2017dynamic} investigated a similar problem as ours by learning edge-conditioned feature weight matrix from edge features using a separate filter-generating network \cite{de2016dynamic}, while \cite{simonovsky2017dynamic}'s application is on point cloud classification. 
There are other studies about learning on graph data such as \cite{dai2016discriminative} that proposed a kernel embedding methods on feature space for graph-structured data. Another similar work is \cite{grover2016node2vec}, but their models do not fall into the kingdom of feed-forward CNN analogs on graphs.

For chemical compounds, naturally modeled as graphs, \cite{duvenaud2015convolutional,wallach2015atomnet,wu2017moleculenet} made several successful trials of applying neural networks to learn representations for predictive tasks, which were usually tackled by handcrafted features \cite{mayr2016deeptox} or hashing \cite{weiss2009spectral}. Whereas, due to the constraints of spatial convolution, their models failed to make full use of the atom-connectivities, which are more than bond features by \emph{Rdkit} \cite{landrum2013rdkit}. More recent explorations on progressive network, multi-task learning and low-shot or one-shot learning have been accomplished \cite{altae2016low,gomes2017atomic}. Lastly, Deepchem \footnote[1]{\url{https://github.com/deepchem/deepchem}} is an outstanding open-source chem-informatics/machine learning benchmark. Our codes and demos were built and tested upon it.

\section{Method} \label{sec:gcn_ll}
\subsection{Spatial v.s Spectral Convolution}
For constructing convolution operators on graph-structured data, there exist two major approaches: spatial construction and spectral construction. As implied by the name of the two, they separately manipulate spatial and spectrum domain of graph signals. Particularly, spatial convolution purely uses neighborhood information in terms of graph adjacency matrix $A_k$ or similarity matrix $W_k$. More formally, if at $k$th layer the input data $x_k \in \mathbb{R}^{n_{k-1}\times d_{k-1}}$, its output $x_{k+1}$ is formulated as \cite{bruna2013invariant,wu2017moleculenet}:\begin{equation}
x_{k+1, j} = \sigma\big( \sum_{i=1}^{f_{k-1}} F_{k,i,j} x_{k,i} \big) \quad j=1,\cdots,f_k \label{eq:spatial}
\end{equation}
where $F_{k,i,j}$ is a $d_{k-1}\times d_{k-1}$ matrix that linearly maps each input feature dimension to output features and possibly $f_k\neq f_{k-1}$. Nonzero entries of $F_{k,i,j}$ are where two nodes connected. Apparently, this model is hard to induce weights shared across spatial domain. The convolution of this type reduces to an analog of fully connected layer with sparse regularization given by $A_k$ on weight matrix $F_{k,i,j}$. See Fig. \ref{fig:spatial} for explicit demonstrations of spatial graph convolution and graph max pooling layers.

Compared to spatial construction, spectral graph theory empowers us to build convolution kernel on spectrum domain which is more compact and the spatial locality of kernel is supported by the smoothness of spectrum multipliers. The baseline approach is built upon [Eq(3), \cite{defferrard2016convolutional}] which extended the one-hop spatial kernels \cite{bruna2013spectral} to the kernels that allow $K$-hop connectivities. According to graph Fourier transform \cite{defferrard2016convolutional}, if $U$ is graph Fourier basis of $L$:\begin{equation}
x_{k+1} = \sigma\big( g_{\theta}(L^K) x_{k} \big) = \sigma\big( U g_{\theta}(\Lambda^K) U^T x_{k} \big), \label{eq:kthspectral}
\end{equation}
where $diag(\Lambda)$ is $\mathcal{O}(N)$ frequencies of Laplacian $L$. The Eq.(\ref{eq:kthspectral}) brings us an elastic kernel that allows any pair of nodes with shortest path distance $d_{\mathcal{G}}<K$ to squeeze in. Of course, the far-away connectivity means less similarity and will be assigned less importance by $g_{\theta}(\Lambda^K)=\sum_{k=0}^{K-1}\theta_k \Lambda^k$.

\textbf{Recursively fast filtering.} Evaluating Eq.(\ref{eq:kthspectral}) is expensive due to dense matrix multiplication with $U$. For instead, $\theta$ and $g_{\theta}(\star)$ were approximated by Chebyshev coefficients and polynomial functions. The computation of $g_{\theta}(L^K)x$ were replaced by recursive function $T_k(x)=2x T_{k-1}(x) - T_{k-2}(x)$ with $T_0=1$ and $T_1=x$. Then the $K$-hop kernel becomes $g_{\theta}(\Lambda) = \sum_{k=0}^{K-1}\theta_k T_k(\tilde{L})$ still parameterized by vector $\theta$ of size $K$. Consequently, the entire cost was reduced to $\mathcal{O}(K)$ from $\mathcal{O}(N^2)$ because of the natural sparsity of $L$ \cite{defferrard2016convolutional}. 

\textbf{Re-parameterization on feature domain.} One major idea for graph CNN is to exactly reconstruct classical CNN on graphs. This way is tough, because regularly shaped kernel is impossible on graphs. \cite{bruna2013invariant,duvenaud2015convolutional} simply bypass building kernels on spatial domain, but give feature transformation conditioned on edge \emph{distance} \cite{bruna2013invariant} or even node degree \cite{duvenaud2015convolutional}. Spectral kernel Eq.(\ref{eq:kthspectral}) is a promising attempt. But it distributes weights in spatial domain similarly in concentric zone model, which is still not as flexible as convolution kernel on grid. Besides, for convolution layer of classical CNNs, outputted activations combine filtered signals from all feature maps in which separate kernels work independently. In other words, they do not only sum up features from their spatial neighbors, but also mine relationships with other feature dimensions. To mimic the classical CNNs, we re-parameterize output of Eq.(\ref{eq:kthspectral}) by a feature domain transformation matrix $W_k \in \mathbb{R}^{f_{k}\times f_{k-1}}$ and a bias $b_k \in \mathbb{R}^{f_k\times 1}$. Intuitively, we divide the operations of classical CNNs on both spatial and feature domain into \emph{two} consecutive stages: 1) compute $g_{\theta}(L^K)$ kernel with $x_k$; 2) linearly maps $f_{k-1}$ features to another $f_{k}$ features. The layer after re-parameterization is as below:\begin{equation}
x_{k+1} = \sigma\bigg( \big(\sum_{k=0}^{K-1}\theta_k T_k(\tilde{L}, x_k)\big)W_k + b_k \bigg).  \label{eq:kthspectral_rp}
\end{equation}

\subsection{Graph CNN with Laplacian Learning}
The state-of-the-art methods on graph convolution neural networks all utilize graph Laplacian matrix in some way. Normalized graph Laplacian is more often used. Given the adjacency matrix $A$ and the degree matrix $D$ for graph $\mathcal{G}=(V, E)$, the graph Laplacian matrix :\begin{equation}
L = I - D^{-1/2} A D^{-1/2} \label{eq:laplacian}
\end{equation}
As we know, $L$ defines both node-wise connectivity and degree of vertices. Some types of data have inherent graph structure, such as chemical molecular data. Each molecule is a graph with atoms as vertices and bonds as edges. Those chemical bonds could be verified by experiments and even visible in some cases. But, most of data do not have graph structure given, so we have to construct graphs before feed them to our deep nets. Besides above two cases, the most likely case is that the inherent graphs can not sufficiently express all of the meaningful node-wise connectivities. For example, \cite{mayr2016deeptox} proposed to predict the toxicity of drugs by learning representations of toxic sub-structures from labeled molecular samples. The graph directly given by SMILES \cite{weininger1988smiles} sequence does not tell anything about the toxicity. The model has to learn the atom connectivity and to form sub-structures most related to toxicity. The discovered toxic sub-structure may happen to be of the bonds, e.g. Benzene ring, or not at all.
Given this, the next question becomes what defines a particularly good distance metric that best describes those hidden connectivities driven by learning tasks.

\begin{figure}[h]
	\centering
	\includegraphics[width=0.5\textheight]{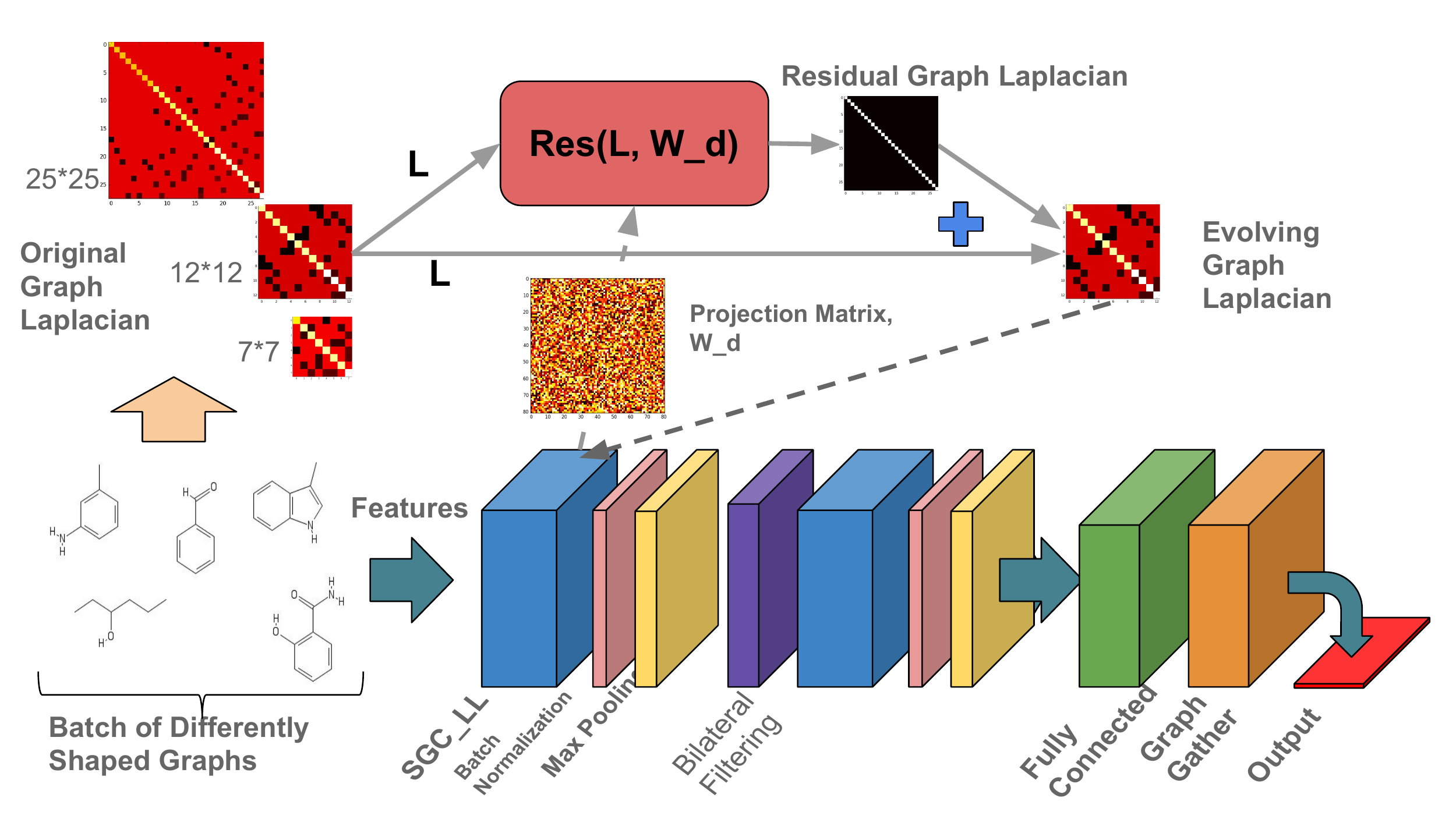}
	\caption{The composition of Evolving Graph Convolution Neural Network (EGCN) and the residual Laplacian learning scheme of SGC-LL layer. The evolving graph Laplacian $L_e = Res(L, W_d) + L$.  ($\sigma=1$) Projection weight $W_d$ in distance metric get updated by SGC-LL in back-propagation. In feed-forward, use $L_e$ and $x$ to compute spectral convolution (Eq.(\ref{eq:kthspectral_rp})). Better viewed in color print.} \label{fig:sgc_ll}
\end{figure}

\textbf{Supervised Metric Learning.} In articles of metric learning, the algorithms were divided as supervised and unsupervised learning for metrics \cite{wang2015survey}. The unsupervised metric selection picks the metric that works best for clustering data samples. The optimal metric should minimize the intra-cluster distances and also maximize the inter-cluster distances. For datasets come with labels, the quality of metric is determined by the learning loss. Parameterized as part of learning model, the metric converges to the optimal when the learning curve remains stable. \emph{Generalized Mahalanobis distance} measures the distance between samples $x$ and $y$ by:\begin{equation}
\mathbb{D}(x, y) = \sqrt{(x-y)^T M (x-y)}. \label{eq:m_dist}
\end{equation}
If $M=I$, Eq.(\ref{eq:m_dist}) reduces to Euclidean distance. In proposed EGCN, the symmetric positive semi-definite matrix $M=W_d W_d^T$ is the trainable weight of SGC-LL layer. The $W_d^k\in \mathbb{R}^{f_k\times f_k}$ works as a transform basis to some domain in which we measure the Euclidean distance between $x$ and $y$. Then, we use that distance to calculate the Gaussian kernel : $\mathbb{G}(x,y) = \exp(-\mathbb{D}(x, y)/(2\sigma^2))$.
In our case, the optimal transformation matrix $\hat{W}_d^k$ will be found by the one who is able to generate the graphs $\hat{L}$ that best fit our learning tasks. Although the distance formulation Eq.(\ref{eq:m_dist}) seems trivial, it is cheap to compute gradient w.r.t $W_d^k$ in back-propagation, which is the main source of computations in DNN.

\textbf{Learning Residual Graph Laplacian} As we discussed above, to discover the hidden correlations between nodes in graph, we introduce a parameterized distance Eq.(\ref{eq:m_dist}) to update the Gaussian similarity matrix $S$ ($A$, adjacency matrix after thresholding), and then use updated $A$ to compute normalized graph Laplacian $L$ (Eq.(\ref{eq:laplacian})). Due to the distance parameter $W_d^k$ is randomly initialized, so it may take long before the model to converge. To accelerate the convergence and increase the stability of our model, we announce a reasonable assumption that the optimal graph Laplacian $\hat{L}=\mathcal{F}(L_{orig}, W_d^k)$ is a small shifting from the original graph Laplacian $L_{orig}$, in other words the original graph Laplacian has disclosed a large amount of helpful graph structural information. Consequently, instead of directly learning $\mathcal{F}(L_{orig}, W_d^k)$, we learn the residual graph Laplacian: $Res(L_{orig}, W_d^k)$, so we have :\begin{equation}
Res(L_{orig}, W_d^k) = \mathcal{F} - \sigma L_{orig}. 
\end{equation}
The proposed Laplacian-Learning boosted spectral graph convolution layer is fed by mini-batch of arbitrarily shaped graphs, it performs convolution on spectrum domain with $K$-hop elastic kernel of $\mathcal{O}(K+f_{k}f_{k-1} + f_{k-1}^2)$ training parameters. $\mathcal{O}(f_{k-1}^2)$ is for the weights $W_d^k$ in distance metric. In Fig. \ref{fig:sgc_ll}, the network consists of two SGC-LL layers, in which the two sets of graph Laplacian $L_k$ will be updated independently and will probably diverge because the input $x_k$ is different, and the two layers worked on different feature maps.

In Section. \ref{sec:exp}, our experimental results on multiple datasets indicate that for the data with inherent graphs, e.g. drug data given as SMILES sequences, the original Laplacian is quite close to the optimal one. However, those small updates on graph connectivity within 20 epochs significantly raise the performance of model. $L_{orig}$ plays a role similar to $l_1$ regularization on $L$ and its weight is balanced by $\sigma$. For those datasets without given graphs, we could use clustering algorithms, e.g. $k$-nearest neighbor, spectral clustering \cite{ng2001spectral}, to construct graphs in unsupervised fashion. Then using them as initialization of the network is better than purely randomized weights initializer. See Fig. \ref{fig:sgc_ll} for details of SGC-LL layer and the residual graph Laplacian learning procedures in this layer.

\section{Experiments}
\label{sec:exp}
\textbf{Network Configuration of EGCN.} The proposed network is named as \emph{evolving} graph convolution networks (EGCN), because it allows graph structure evolves according to the context of learning task. Besides SGC-LL, it has graph max pooling layer and gathering layer \cite{gomes2017atomic}. The max pooling on graph was performed feature by feature. For each node $v$, the operator replaces the $j$th feature of node $v$ with the maximum among the original values from his neighbors $N(v)$ and himself: $\hat{v}_j = \max(\{v_j, i_j \forall i \in N(v)\})$. In graph gather layer, it simply sums up the feature vectors of all nodes and output it as representation of the graph, so we can use it to do graph-level regression or classification. The motivation of embedding a bilateral filter in EGCN is against over-fitting \cite{gadde2016superpixel}. The evolving graph Laplacian definitely adapts the model to better fit the training data, but, at the risk of over-fitting.  To prevent over-fitting, we introduce a revised bilateral filtering layer to regularize activation of SGC-LL layer by augmenting the spatial locality of $L$. We also introduced batch normalization layers to accelerate the training \cite{ioffe2015batch}.

\textbf{Batch Training of Non-uniformly Shaped Samples.} One of the greatest challenges for graph CNN is the different shapes of training graphs: 1) Raises the difficulty of designing kernels, because the invariance of kernel on graphs is not satisfied and the node indexing sometimes matters; 2) Sometimes resizing (clustering) \cite{bruna2013invariant} is not reasonable for some types of graph like molecular data: it will lose significant atoms along with its features, if perform graph coarsening or pooling; 3) Most of deep learning APIs do not support the training inputs of varying shapes in batch-mode \footnote[1]{I see some new workout released by Google's Tensorflow \cite{looks2017deep}, but due to time constraint, we do not try to move our code to that frameworks.}. In this work, we bypassed the tensor shape constraint of Tensorflow by heavy usage of $tf.pad$ and $tf.slice$. Samples has different number of nodes, so their graph Laplacians definitely differ, but they share all the model parameters.  In the experiments, we almost reused the parameter setup for all datasets. Batch size is 256. The optimizer is \emph{Adam} with exponential decayed (0.9 every 50 iterations) learning rate begins with 0.005. The maximum epoch is 50. We extracted 75 node features and 6 edge features. 

\begin{figure}[h]
	\centering
	\includegraphics[width=0.6\textheight]{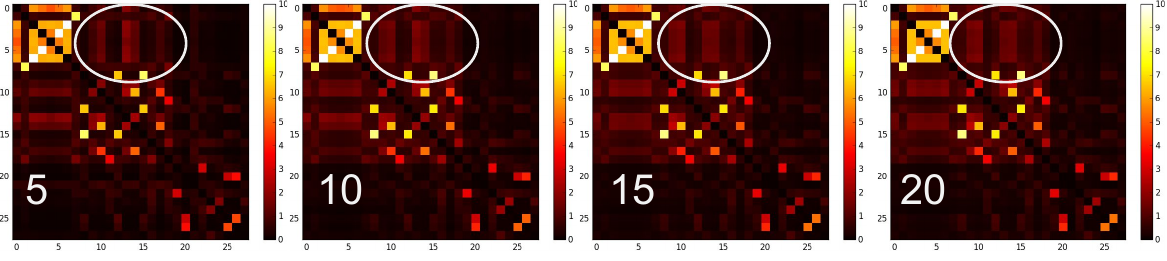}
	\caption{Example of residual graph Laplacian learning. Four shots of evolving graph similarity matrix $S$ (shown in heat maps) recorded at 5th, 10th,15th and 20th epoch separately. The displayed $28\times 28$ $S$ is from compound \emph{Benfuracarb} that has 28 atoms. } \label{fig:LL}
\end{figure}
%

\begin{figure}
	\topinset{\bfseries(a)}{\includegraphics[width=0.5\textwidth]{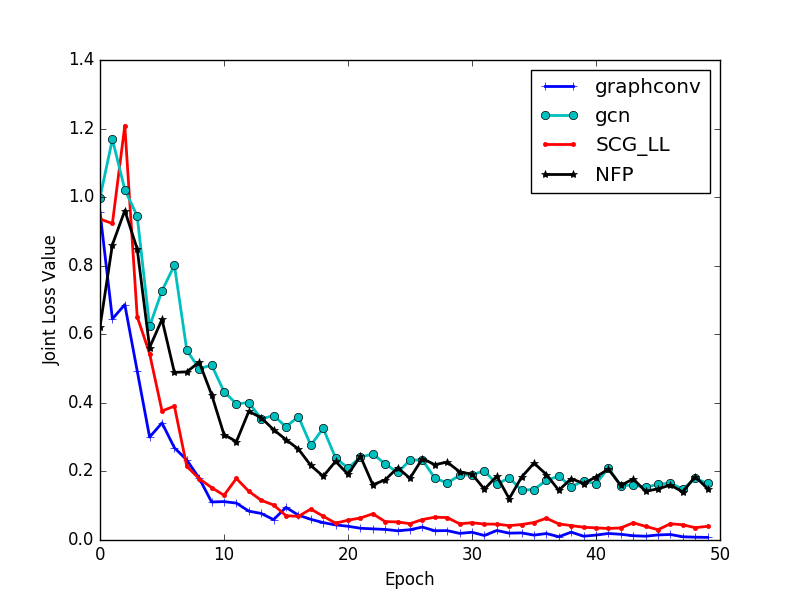}}{0.25in}{.4in}
	\topinset{\bfseries(b)}{\includegraphics[width=0.5\textwidth]{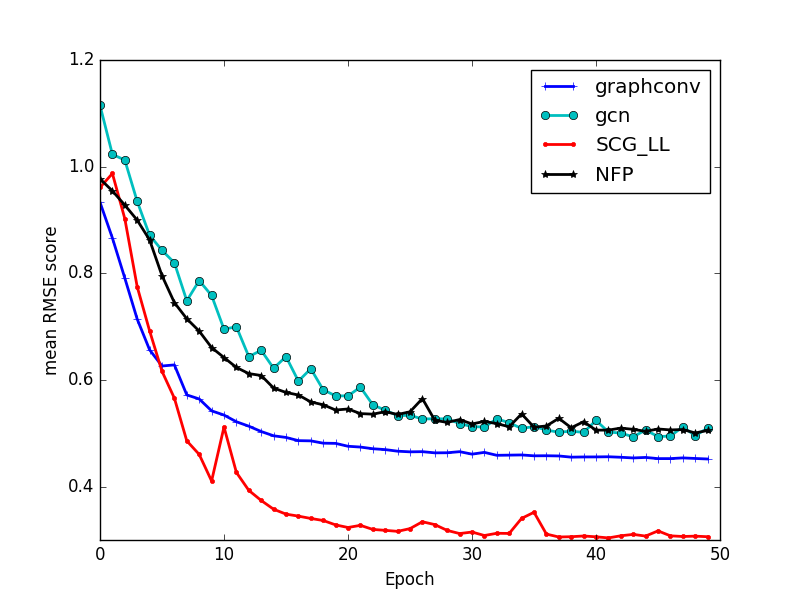}}{0.25in}{.4in}
	\caption{ The weighted $l_2$ loss curve (left) during training and mean RMSE curve (right) from the epoch 0$\sim$50 on validation set of Delaney solubility dataset. Red lines were generated by our "SGC-LL". Compared benchmarks: graphconv \cite{bruna2013spectral}, gcn \cite{defferrard2016convolutional}, NFP \cite{duvenaud2015convolutional}. Better viewed in color print.}\label{fig:curve}
\end{figure}

\subsection{Performance boosted by SGC-LL Layer}\label{sec:LL}
 The experiment demonstrates the close correlation between evolving graph Laplacian and model fitting. Fig. \ref{fig:LL} shows the 4 heat maps of graph similarity matrix $S$, used to compute evolving graph Laplacian, at the second SGC-LL layer. As shown in Fig. \ref{fig:curve}, the weighted $l_2$ loss dropped quickly during the epoch 5$\sim$20, so did the mean RMSE score. In the meanwhile, the graph Laplacians keep evolving according to gradient back-propagated from next layer. The white circles mark one of the major region of intensities on $S$ that changed significantly during the epoch 5$\sim$20. The connections between some pairs of node were reinforced (get lighter), while others got weakened (go darker). Besides, in Fig. \ref{fig:curve}, the two figures show that the EGCN network equipped with proposed SGC-LL layer (red line) has overwhelmingly better performance in both convergence speed and predictive accuracy. We attribute this improvement to the supervised residual graph Laplacian learning scheme during training. The evolving graph Laplacians, used in spectral graph convolution, fit the data better than the fixed graph Laplacian \cite{defferrard2016convolutional,kipf2016semi}.

\begin{table}[t]
	\caption{mean RMSE Scores on Delaney, Az-logD and NCI Datasets}
	\label{tab:regression-table}
	\centering
	\begin{tabular}{l|c|c|c }
		\hline 
		Datasets    & Delaney solubility & Az-logD & NCI \\
		\hline \hline
		G-CNN \cite{bruna2013spectral}   &  0.42225 $\pm$ 8.38$\mathrm{e}{-4}$  &  0.75160 $\pm$ 8.42$\mathrm{e}{-3}$ & 0.86958 $\pm$ 3.55$\mathrm{e}{-4}$  \\ 
		NFP \cite{duvenaud2015convolutional}   &  0.49546 $\pm$ 2.30$\mathrm{e}{-3}$ & 0.95971 $\pm$ 5.70$\mathrm{e}{-4}$ & 0.87482 $\pm$ 7.50$\mathrm{e}{-4}$  \\ 
		GCN  \cite{defferrard2016convolutional}     &   0.46647 $\pm$ 7.07$\mathrm{e}{-3}$  & 1.04595 $\pm$ 3.92$\mathrm{e}{-3}$ & 0.87175 $\pm$ 4.14$\mathrm{e}{-3}$  \\  
		SGC-LL      & \textbf{0.30607} $\pm$ 5.34$\mathrm{e}{-4}$ & \textbf{0.73624} $\pm$ 3.54$\mathrm{e}{-3}$ &  \textbf{0.86474} $\pm$ 4.67$\mathrm{e}{-3}$  \\
		\hline
	\end{tabular}
\end{table}
\begin{table}[t]
	\caption{Task-averaged ROC-AUC Scores on Tox21, ClinTox, Sider $\&$ Toxcast Datasets} \label{tab:classification}
	\centering
	\begin{tabular}{l|c|c|c|c|c|c|c|c}
		\hline
		 Datasets    &\multicolumn{2}{|c|}{Tox21} & \multicolumn{2}{|c}{ClinTox} & \multicolumn{2}{|c}{Sider} & \multicolumn{2}{|c}{Toxcast}\\
		     \hline
		     &  Valid   & Test  & Valid & Test  & Valid & Test & Valid & Test  \\
		\hline \hline 
		G-CNN \cite{bruna2013spectral}     & 0.7105  & 0.7023 &  0.7896 & 0.7069 & 0.5806 &  0.5642 & 0.6497 & 0.6496\\ 
		NFP \cite{duvenaud2015convolutional}  & 0.7502  & 0.7341 & 0.7356 & 0.7469 &  0.6049 & 0.5525 & 0.6561 & 0.6384  \\
		GCN  \cite{defferrard2016convolutional}      &  0.7540  & 0.7481 & 0.8303 & 0.7573 & 0.6085 & 0.5914 & 0.6914 & 0.6739 \\  
		SGC-LL       &  \textbf{0.7947}    & \textbf{0.8016} &  \textbf{0.9267}  & \textbf{0.8678} & \textbf{0.6112} & \textbf{0.5921} & \textbf{0.7227} & \textbf{0.7033}  \\
		\hline
	\end{tabular}
\end{table}

\subsection{Prediction on Chemical Molecular Datasets}
Delaney Dataset \footnote[1]{Delaney Dataset: \url{http://pubs.acs.org/doi/abs/10.1021/ci034243x}} contains aequeous solubility data for 1,144 low molecular weight compounds. The complexest compound in the dataset has 492 atoms, while the smallest one only consists of 3 atoms. For organic compound, we set the maximum degree of node as 10. NCI chemical compound database \footnote[2]{NIH-NCI: \url{https://cactus.nci.nih.gov/download/nci/}} has around 20,000 training compound samples and 60 prediction tasks from drug reaction experiments to clinical pharmacology studies. At last, Az-logD dataset from ADME \cite{vugmeyster2012absorption} is a set of compounds and their logD measurements correlated to permeability. The presented task-averaged RMSE scores and standard deviations were obtained after 5-fold cross-validation.

To demonstrate our advantage, we compared it with three state-of-the-art graph CNN benchmarks: the pioneering graph CNN (G-CNN) \cite{bruna2013spectral}, its spectral domain extension to $k$-hop (GCN) \cite{defferrard2016convolutional} and neural fingerprint (NFP) \cite{duvenaud2015convolutional}. In Table. \ref{tab:regression-table}, our network reduced the mean RMSE by 31$\%$ -40$\%$ on Delaney dataset, averagely 15$\%$ on Az-logD and 2$\sim$4$\%$ on testing set of NCI. The improvements come from the more meaningful representations extracted by SGC-LL layer. First, the $k$-hop kernel on spatial domain via Eq.(\ref{eq:kthspectral}) used to be impossible \cite{bruna2013spectral,duvenaud2015convolutional}, then re-parameterization offers feature domain filter mappings that was absent in \cite{defferrard2016convolutional}. Besides, our residual Laplacian learning and updating scheme does learn a better graph structure that optimally fits the learning tasks while training, which makes more sense than graphs constructed by unsupervised clustering \cite{gadde2016superpixel} or separate training networks \cite{henaff2015deep,simonovsky2017dynamic}.

\subsection{Multi-task Classification on Pharmacological Datasets}
Tox21 Dataset \cite{mayr2016deeptox} we used contains 7,950 chemical compounds. It has 12 classification tasks for different essays of toxicity, however, not every sample contains all 12 labels. For those with missing labels, we excluded them when computing losses, but still kept them in train dataset. ClinTox is a public dataset of 1451 chemical compounds for clinical toxicological study together with labels for 2 tasks. Sider \footnote[3]{Sider Data Web: \url{http://sideeffects.embl.de/}} database records 1392 drugs and their 27 different side effect or adverse reaction. Toxcast is another toxicological research database that has 8,571 SMILES samples and the database has labels for 617 predictive tasks. For $N$-task prediction, the network graph model will become an analog of K-ary tree with $N$ leaf nodes, each of which is actually a fully connected layer followed by logistic regression to generate scores for each task \cite{mayr2016deeptox}. The displayed scores were averaged over all tasks at Table. \ref{tab:classification}. Obviously, our method greatly raises classification accuracy on both small and large datasets, and even 5$\%$ on average for 617 tasks on Toxcast dataset.

\section{Conclusions}
We proposed a new spectral graph convolution layer that learns the residual graph Laplacians via learning optimal metric weights. The proposed EGCN can be fed by a batch of arbitrarily shaped samples on graph. For each sample, the network can individually learn the graph structure that optimally expresses the hidden node-wise connectivity. The training in a supervised fashion was driven by context of learning tasks. The extensive experiments show that our evolving graph CNN outperforms the state-of-the-arts on multiple datasets. In future, we plan to design a real spatial kernel of elastic kernel on graphs. Second, the implementation of SGC-LL need to be remodeled and hopefully get accelerated. Another interesting work is to extend graph CNNs to applications such as natural language understanding and user-behavior prediction on social networks.

\small
\bibliographystyle{IEEEtran}
\bibliography{IEEEabrv,ref}
\end{document}